\documentclass[sigconf]{acmart}

\usepackage{float}
\usepackage{enumitem}
\usepackage[noend]{algpseudocode}
\usepackage{algorithm}
\usepackage{amsmath}
\setlist[itemize]{noitemsep, topsep=0pt}
\AtBeginDocument{%
    \providecommand\BibTeX{{%
    \normalfont B\kern-0.5em{\scshape i\kern-0.25em b}\kern-0.8em\TeX}}}

\setcopyright{acmcopyright}

\setcopyright{none}
\acmConference[]{PREPRINT}{April}{2024}
\acmDOI{10.48550/arXiv.2402.01275} 
\acmISBN{}

\newcommand\ourMethodAccronyme{PT-ME}

\definecolor{light-gray}{gray}{0.92}
\definecolor{dark-gray}{gray}{0.85}
\definecolor{light-blue}{RGB}{207, 226, 243}
\definecolor{light-green}{RGB}{217, 234, 211}
\definecolor{light-orange}{RGB}{252, 229, 205}

\algrenewcommand{\algorithmiccomment}[1]{\hfill\colorbox{light-gray}{$\triangleright$ #1}}
\newcommand\highlightSection[1]{\colorbox{dark-gray}{#1}}
\newcommand\HLreg[1]{\colorbox{light-blue}{#1}}
\newcommand\HLpar[1]{\colorbox{light-green}{#1}}

\begin{document}

\title{Parametric-Task MAP-Elites}

\author{Timothée Anne}
\orcid{0000-0002-4805-0213}
\author{Jean-Baptiste Mouret}
\orcid{0000-0002-2513-027X}
\email{jean-baptiste.mouret@inria.fr}
\affiliation{%
    \institution{Université de Lorraine, CNRS, Inria}
    \city{F-54000 Nancy}
    \country{France}
}

\renewcommand{\shortauthors}{Anne and Mouret}

\begin{abstract}
    Optimizing a set of functions simultaneously by leveraging their similarity is called multi-task optimization. Current black-box multi-task algorithms only solve a finite set of tasks, even when the tasks originate from a continuous space. In this paper, we introduce Parametric-Task MAP-Elites (PT-ME), a new black-box algorithm for continuous multi-task optimization problems. This algorithm (1) solves a new task at each iteration, effectively covering the continuous space, and (2) exploits a new variation operator based on local linear regression. The resulting dataset of solutions makes it possible to create a function that maps any task parameter to its optimal solution. We show that PT-ME outperforms all baselines, including the deep reinforcement learning algorithm PPO on two parametric-task toy problems and a robotic problem in simulation. 
\end{abstract}



\begin{teaserfigure}
    \includegraphics[width=\textwidth]{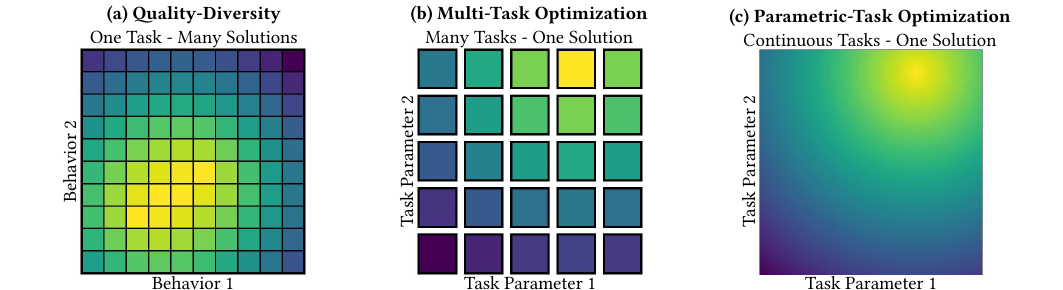}
    \caption{(a) Quality-Diversity is the problem of finding high-performing solutions with diverse behaviors. (b) Multi-task optimization is the problem of finding the optimum solutions for a finite set of tasks, each often characterized by a task parameter (or descriptor). (c) In this paper, we propose to extend the multi-task optimization problem to a continuous parametrization, which we call Parametric-Task Optimization. The goal is to be able to return the optimal solution for any task parameter. }\label{fig:_teaser}
\end{teaserfigure}

\maketitle

\section{Introduction}

Optimization is ubiquitous in today's world~\cite{Optimization}. While many problems require optimizing one cost function, many involve a set of related functions, each representing a different ``task''. For example, in robotics, it can be useful to learn walking gaits for various legged-robot morphologies~\cite{Multi-Task_MAP-Elites}, with each task corresponding to maximizing the walking speed for one morphology. Instead of solving each optimization individually, solving all the tasks simultaneously should decrease the total computation cost. Another illustration can be found in machine learning, where optimizing the hyperparameters of an algorithm for specific datasets can make a difference; here, each task is the loss function minimization for each dataset~\cite{EMT_aplication_hyperparameter_GP}. 

Multi-task problems are encountered in many fields such as robotics~\cite{Multi-Task_MAP-Elites, goal_babbling_baranes}, embodied intelligence~\cite{Multi-Task_MAP-Elites, EMT_planar_robotic_arm_anomaly_detection, EMT_double_pole_neuroevolution}, data science for symbolic regression of different real-dataset using genetic programming~\cite{EMT_symbolic_regression_with_GP}, multi-software testing using multifactorial evolutionary algorithm~\cite{EMT_software_testing_multifactorial}, logistics planning for different vehicle routing problems~\cite{EMT_logistics_planning_combinatorial_optimzation}, complex design for cars~\cite{EMT_complex_design_cars}, packaging~\cite{EMT_complex_design_packaging}, photovoltaic models~\cite{EMT_complex_design_photovoltaic_models}, or electric power dispatch~\cite{EMT_complex_design_electric_power_dispatch}.

To our knowledge, all black-box multi-task algorithms solve a finite set of tasks, whereas many problem specifications are continuous. This means that continuous task spaces have to be discretized. For example, in robotics, the morphology is characterized by continuous length of limbs and joint bounds, and any discretization is arbitrary. When the cost function is differentiable, a few algorithms exist (i.e., Parametric programming~\cite{multiparametric-programming}) to solve the continuous multi-task problem, which we call the parametric-task optimization problem, but we are not aware of any black-box algorithm to do so.

The contribution of this paper is a black-box parametric-task algorithm called Parametric-Task MAP-Elites (PT-ME) inspired by Multi-Task MAP-Elites (MT-ME)~\cite{Multi-Task_MAP-Elites}, an elitist multi-task algorithm. 
Our main idea is to design an algorithm that solves as many tasks as possible within its allocated budget, asymptotically solving the parametric-task optimization problem by filling the task space with high-quality solutions. The resulting dense dataset can then be distilled, for instance, with a neural network, making it possible to generalize and obtain a high-quality solution for any task.  

PT-ME's key features are the following:
\begin{itemize}
    \item it samples a new task at each iteration, meaning that the more evaluation budget it has, the more tasks it will solve;
    \item its selection pressure is de-correlated from the number of tasks it solves, meaning that its efficiency does not decrease with the number of tasks to solve;
    \item it uses a specially designed variation operator that exploits the multi-task problem structure to improve the performance;
    \item it distillates those solutions into a continuous function approximation, i.e., a neural network, effectively solving the parametric-task optimization problem.
\end{itemize}

We evaluate our method and its ablations extensively on two parametric-task optimization toy problems, 10-DoF Arm (used in~\cite{Multi-Task_MAP-Elites, EMT_planar_robotic_arm_anomaly_detection, many-tasks_EME-BI}) and Archery (a more challenging problem with sparse rewards), and a more realistic parametric-task optimization, Door-Pulling, consisting of a humanoid robot pulling a door open. We compare against several baselines and show that our method effectively solves the parametric-task optimization problem while being more efficient than all baselines and ablations.

\section{Problem formulation}
\label{sec:_formulation}

The goal is to find a function $G$ that outputs the optimal solution for each task parameter $\theta$. More formally:
\begin{equation}
    \forall \theta \in \Theta, G(\theta) = x^*_{\theta} = \underset{x \in \mathcal{X}}{\operatorname{argmax}}  (f(x,\theta))
\end{equation}     
where $\mathcal{X}$ is the solution space, $\Theta$ the task parameter space, $f: \mathcal{X}\times\Theta \rightarrow \mathbb{R}$ the function to optimize (also called fitness function), and $G: \Theta \rightarrow \mathcal{X}$ a function that outputs the optimal solution $x^*_{\theta}$ maximizing the fitness function for the task parameter $\theta$. 
The main difference with multi-task optimization is that the task space $\Theta$ is continuous and not a finite set of tasks. 

For ease of use in the remaining of the paper, we set the solution space $\mathcal{X} = {[0,1]}^{d_x}$ where $d_x$ is its dimension and the parameter space $\Theta = {[0,1]}^{d_{\theta}}$ where $d_{\theta}$ is its dimension.

\section{Related Work}

\subsection{Parametric Programming} 

Optimizing a function parameterized with discrete or continuous parameters is studied in mathematics and is called parametric programming (or multiparametric programming for multiple parameters)~\cite{pLP, mpLP, mpQP, mpNLP, multiparametric-programming}. The main algorithms divide the parameter space into critical regions and solve the optimization problem on each critical region using classical mathematical tools, i.e., Lagrange multipliers and Karush-Kuhn-Tucker conditions. There are three families of methods: multiparametric linear programming (mpLP)~\cite{mpLP}, multiparametric quadratic programming problems (mpQP)~\cite{mpQP}, and multiparametric nonlinear programming~\cite{mpNLP}, which approximates nonlinear functions. 

One application is to use parametric programming to precompute the optimal solutions for all parameter space and then query those solutions online. The reformulation of model predictive control problems with a quadratic cost function into mpQP~\cite{multiparametric-programming} has led to MPC on chip (or explicit MPC) with applications, for example, to control chemical plants~\cite{MPC_on_chip}.  

All parametric programming algorithms assume that the function to optimize is known. While there are many situations where that information is available, for many real-world applications (e.g., complex simulations or real systems), the functions to optimize are non-linear, non-convex, non-differentiable, or even black-box, making those methods inapplicable. 

\subsection{Multi-Task Optimization}

In recent years, multi-task optimization has seen a surge of interest in the evolutionary computation community~\cite{Multifactorial_evolution, multi_task_survey, EMT_6_applications}. We can decompose it into two families: implicit transfer of solutions (i.e., one solution space for all tasks) and explicit transfer of solutions (i.e., different solution spaces). Implicit transfer is more straightforward as the algorithm evolves only one population of solutions for all tasks but only applies to homogenous tasks. Explicit transfer concerns problems with heterogeneous tasks where knowing which information to share between tasks is not trivial. Proposed methods use probabilistic search distribution~\cite{EMT_probabilistic_search_distribution}, search direction vectors~\cite{EMT_search_direction_vectors}, higher-order heuristics~\cite{EMT_higer-order_heuristics}, or surrogate models~\cite{EMT_surogate_model}.

Those methods have applications in many fields but rarely consider more than a handful of tasks simultaneously (i.e., between 2 and 10): $3$ morphologies for robot controller using neuroevolution~\cite{EMT_double_pole_neuroevolution}, $2$ datasets for symbolic regression using genetic programming~\cite{EMT_symbolic_regression_with_GP}, $10$ numerical calculus functions in C for software testing or $3$ car models for design optimization using a multifactorial evolutionary algorithm~\cite{EMT_software_testing_multifactorial, EMT_complex_design_cars}, $2$ auxiliary tasks and one main task (i.e., $3$ tasks) for safe multi-UAV path planning~\cite{EMT_UAV_path_planning} or $2$ bus systems power dispatch optimization using a multi-objective multifactorial evolutionary algorithm~\cite{EMT_complex_design_electric_power_dispatch}, $4$ package delivery problems planning using explicit transfer knowledge~\cite{EMT_logistics_planning_combinatorial_optimzation}, $3$ diode models design using similarity-guided evolutionary multi-task optimization~\cite{EMT_complex_design_photovoltaic_models}.   

Other fields, such as Bayesian optimization~\cite{Bayesian-Optimization_review, Bayesian-Optimization_tuto}, propose multi-task algorithms, for example, to tune machine learning algorithms for several datasets~\cite{Multi_task_bayesian_optimization}. Bayesian optimization is more focused on expensive fitness functions and cannot handle more than a thousand evaluations due to the cubic scaling of the Gaussian processes. 

A handful of works tackle more than a dozen tasks: 30 in~\cite{many-tasks_EBS}, 50 in~\cite{many-tasks_EMaTO-MKT}, 500 in~\cite{EMT_planar_robotic_arm_anomaly_detection}, and $2\,000$ in~\cite{many-tasks_EME-BI}.
In this paper, we take inspiration from MT-ME~\cite{Multi-Task_MAP-Elites}, which scales up to $5\,000$ tasks.

\subsection{Multi-Task MAP-Elites}

Originating from novelty search~\cite{Novelty-Search}, the Multi-dimensional Archive of Phenotypic Elites (MAP-Elites) algorithm~\cite{MAP-Elites} finds a large set of high-quality and diverse solutions to a task (this problem is called Quality-Diversity (QD)). MAP-Elites~\cite{MAP-Elites} has shown promising results in robotics with evolving a repertoire of diverse gaits for legged robots~\cite{MAP_Elites_NaturePaper}, or in video-gaming industries to generate procedural contents like spaceships and game levels~\cite{PCG-QD}. Covariance Matrix Adaptation MAP-Annealing~\cite{CMA-ME, CMA-MAE} mixes MAP-Elites~\cite{MAP-Elites} with the self-adaptation techniques of CMA-ES~\cite{CMA-ES} and outperforms both methods for finding diverse game strategies for the Hearthstone game. None of those variants have been applied to multi-task optimization. This paper takes root from MT-ME~\cite{Multi-Task_MAP-Elites}, which is closer to the original MAP-Elites~\cite{MAP-Elites}.

MT-ME~\cite{Multi-Task_MAP-Elites} has applications in robotics, e.g., instead of optimizing each morphology individually, it simultaneously optimizes walking gaits for $2\,000$ morphologies.
The resulting walking gaits allow robots to travel three times farther than those evolved with CMA-ES~\cite{CMA-ES} run individually with the same budget of iterations. 
It has applications in industrial scenarios, such as optimizing ergonomic behaviors for different worker morphologies in an industrial workstation~\cite{Jacques_paper}. Multi-Task Multi-Behavior MAP-Elites~\cite{MTMB-ME} is an extension that solves the multi-task Quality-Diversity problem and outperforms MAP-Elites~\cite{MAP-Elites} run on each task individually. 

MT-ME~\cite{Multi-Task_MAP-Elites} explicitly discretizes the task space into a finite set of several thousand tasks and updates an archive of the best-known solution for each task, called an elite. At each iteration, it randomly selects two elites from the archive, generates a candidate solution using a variation operator, and evaluates this candidate solution on one of the tasks. If the task has no known solution or the candidate solution has a higher fitness than the current one, it becomes the elite for this task, and the algorithm resumes its main loop.

A first limitation of this discretization is that MT-ME~\cite{Multi-Task_MAP-Elites} can only solve a finite number of tasks even though the original task space is continuous. Discretization does not scale well with dimensions, e.g., a thousand tasks only correspond to $32$ steps per dimension in 2D and $10$ steps per dimension in 3D. Another limitation is that, by being an elitist algorithm, its efficiency in finding solutions is correlated to the selection pressure, which is inversely correlated to the number of cells (i.e., the more cells there are to fill, the easier it is for a bad solution to become an elite, and the longer it takes to fill each cell with a good solution). As the number of cells equals the number of tasks to solve, the more tasks there are to solve, the slower the algorithm is to solve them. 

\subsection{Reinforcement Learning}

As stated Sec.~\ref{sec:_formulation}, the end product of parametric-task optimization algorithms should be a function $G: \Theta \rightarrow \mathcal{X}$ taking a task parameter $\theta$ as input and outputting the corresponding optimal solution $x^*_{\theta}$. This function is similar to the action policy in Deep Reinforcement Learning (DRL)~\cite{DRL_survey, DRL_intro}, $\Pi: S \rightarrow A$ taking the current state descriptor $s\in S$ as input and outputting the corresponding optimal action $a \in A$ with regards to a long-term reward $r\in\mathbb{R}$. We can see the similarity between the task parameter $\theta$ and the state $s$, the solution $x$ and the action $a$, and the fitness $f$ and the reward $r$. The main difference is that parametric-task optimization algorithms optimize the fitness value directly. In contrast, DRL algorithms solve a more complex problem as they optimize the action for a long-term reward and must solve the credit assignment problem. Nevertheless, a DRL algorithm applied with a one-step horizon solves a parametric-task optimization problem. We make this parallel to highlight that, like DRL today, parametric-task optimization could have applications in many fields. Proximal Policy Optimization (PPO)~\cite{PPO} is a mature and widely used~\cite{PPO_Supply_Chain_Management, PPO_aerospace, PPO_finance, PPO_energy, PPO_games, PPO_autonomous_driving, PPO_robotics} DRL algorithm, that we think to be a strong baseline. 

\section{Method}

Parametric-Task MAP-Elites (PT-ME) follows the same principles as MAP-Elites~\cite{MAP-Elites}. After initializing each archive cell with a random elite (Alg.\ref{alg:main_alg}, \texttt{Line}.\ref{lst:line:_init_random_solution}), the main loop consists in generating a new solution candidate, evaluating it for a new task (\texttt{L}.\ref{lst:line:_evaluation}), and updating the archive by replacing the elite with the new solution if its fitness is greater or equal (\texttt{L}.\ref{lst:line:_achive_update_start}-\ref{lst:line:_achive_update_end}). The first difference is that we store all the evaluations (\texttt{L}.\ref{lst:line:_evaluations}) for later use during distillation (Sec.\ref{sec:_distillation}) and evaluation (Sec.\ref{sec:_evaluations_and_measures}).
The second difference is that PT-ME uses MT-ME~\cite{MAP-Elites}'s variation operator for half of the iterations (Sec.\ref{sec:_default_operator}) and uses a new variation operator for the other half (Sec.\ref{sec:_new_operator}). 

\subsection{The Archive of Elites}
\label{sec:_archive}

Our first contribution is de-correlating the number of cells in the archive from the number of tasks solved. Like MT-ME~\cite{Multi-Task_MAP-Elites}, PT-ME divides the task parameter space into regions using CVT~\cite{CVT, fast_cvt} (\texttt{L}.\ref{lst:line:_init_CVT}) and attributes an archive cell for each specific centroid task parameter $\theta_c$ (\texttt{L}.\ref{lst:line:_init_archive_start}-\ref{lst:line:_init_archive_end}). 
Nonetheless, instead of evaluating the candidate solutions only on those tasks, PT-ME samples a task parameter $\theta$ at each iteration (\texttt{L}.\ref{lst:line:_sbx_tasks_cnadidates}-\ref{lst:line:_sbx_tournament} and \texttt{L}.\ref{lst:line:_llr_sample_task}) and assigns it to the cell with the closest centroid $\theta_c$ (\texttt{L}.\ref{lst:line:_sbx_corresponding_cell} and \texttt{L}.\ref{lst:line:_llr_corresponding_cell}). By doing so, PT-ME improves the task parameter space coverage at each iteration and directly benefits from a larger budget of evaluations without reducing the convergence to good solutions in the early stage of the algorithm. 
One potential drawback (that we did not observe in our experiments) is that solutions for close but different tasks can be in competition because they are in the same cell, which means that an ``easy'' task could become the cell's elite and prevent the algorithm from finding solutions to ``harder'' tasks of the same cell. This effect is easily attenuated by increasing the number of cells.

By contrast with MAP-Elites, the number of final solutions in PT-ME will far exceed the number of cells. However, the number of cells still impacts the selection pressure on the elites. At the extremes, with one cell, the selection pressure is maximal as every solution competes with every other, and with an infinite number of cells, the pressure is null as each solution is an elite. In addition, the number of cells impacts the new variation operator (Sec.~\ref{sec:_new_operator}). We found empirically that a value of $N=200$ cells is robust for the parametric-task optimization problems presented Sec.~\ref{sec:_Considered_Problems}.

\subsection{Variation Operator 1: SBX with tournament}
\label{sec:_default_operator}

Like MT-ME~\cite{Multi-Task_MAP-Elites} (but only for half of the iterations), PT-ME biases with a tournament the task selected to evaluate the candidate solution. However, instead of sampling the tasks among a fixed set, it uniformly samples them from the task parameter space. 

More formally, for half of the iterations (\texttt{L}.\ref{lst:line:_sbx_select_parents}-\ref{lst:line:_sbx_generate_offspring}), PT-ME randomly selects two parents, $p_1$ and $p_2$, from the archive of elites (\texttt{L}.\ref{lst:line:_sbx_select_parents}), samples $s$ task parameters $\theta_{1:s}$ (\texttt{L}.\ref{lst:line:_sbx_tasks_cnadidates}), selects the one closest to the task parameter from which $p_1$ was evaluated (\texttt{L}.\ref{lst:line:_sbx_tournament}), and generates an offspring $x$ with SBX~\cite{SBX} (\texttt{L}.\ref{lst:line:_sbx_generate_offspring}). The intuition is that two close tasks have more chances to share close solutions but that always taking the closest tasks would lead to poor exploration. 

Like MT-ME~\cite{Multi-Task_MAP-Elites}, PT-ME chooses the size $s$ of the tournament with a bandit (\texttt{L}.\ref{lst:line:_s_update_start}-\ref{lst:line:_s_update_end}), between a minimal size of $1$ (i.e., no tournament) and maximal size of $500$ (i.e., choosing a close task). PT-ME also uses the UCB1~\cite{UCB1} bandit algorithm, which achieves the optimal regret up to a multiplication constant. The bandit score also corresponds to the candidate solution becoming an elite (\texttt{L}.\ref{lst:line:_bandit_update_start}-\ref{lst:line:_bandit_update_end}). 

We concluded in a preliminary study that the bandit is significantly better than choosing the tournament size $s$ at random but leads to slightly worse performances than fine-tuning it for each problem. We decided to keep it as it removes the need to fine-tune a hyperparameter without a significant loss in performance.

\subsection{Variation Operator 2: Linear Regression}

\label{sec:_new_operator}

SBX~\cite{SBX} does not exploit the multi-task framework because it does not use the fact that we know the task on which the candidate solution will be evaluated. We take inspiration from the tournament, but instead of biasing the task parameter given the parents of the candidate solution, we bias the candidate solution given the task parameter. The idea is to aggregate information from the current archive by building a local model of the function $G:\theta \mapsto x^*_{\theta}$ and use it to estimate a candidate solution. We chose a linear model to ``guess'' the solution for the chosen task given the archive because it is simple and fast.


For half of the iterations (\texttt{L}.\ref{lst:line:_llr_sample_task}-\ref{lst:line:_llr_regression}), PT-ME\@:

\begin{enumerate}
    \item samples a task parameter $\theta$ (\texttt{L}.\ref{lst:line:_llr_sample_task});
    \item looks for the closest centroid $\theta_c$ (\texttt{L}.\ref{lst:line:_llr_corresponding_cell}) using a KDTree~\cite{KDTree} with the SciPy implementation~\cite{SciPy} precomputed at the creation of the archive (\texttt{L}.\ref{lst:line:_init_KDtree});
    \item looks for the adjacent cells $A[\theta_c].{adj}$ (using a Delaunay triangulation~\cite{Delaunay1928} with the SciPy implementation~\cite{SciPy} precomputed at the creation of the archive (\texttt{L}.\ref{lst:line:_init_delaunay})) and extract their task parameters $\boldsymbol{\theta}$ and solutions $\boldsymbol{x}$ (\texttt{L}.\ref{lst:line:_llr_adjacent_parameters}-\ref{lst:line:_llr_adjacent_solutions});
    \item performs a linear least squares, $M = ({\boldsymbol{\theta}}^\mathsf{T} \boldsymbol{\theta})^{-1} \boldsymbol{\theta}^\mathsf{T} \boldsymbol{x}$ (\texttt{L}.\ref{lst:line:_llr_lls});
    \item computes the candidate solution with additional noise (\texttt{L}.\ref{lst:line:_llr_regression}), $x = M \cdot \theta + \sigma_{reg} \cdot \mathcal{N}\bigl( 0, \operatorname{variance}(\boldsymbol{x}) \bigr)$ (where $\sigma_{reg}$ regulates the intensity of the Gaussian noise.)
\end{enumerate}


We use adjacency (i.e., Delaunay triangulation~\cite{Delaunay1928}) instead of distance (e.g., the $k$ closest neighbors or the neighbors closer than $\epsilon$) because it adapts the number of adjacent cells to the dimension, thus removing the need to tune a hyperparameter (i.e., the number of neighbors $k$ or the distance~$\varepsilon$). The regression's noise coefficient $\sigma_{reg}$ balances exploration and exploitation. During preliminary experiments, we found that different problems require different optimal values. We concluded on a standard value of $1$, even though fine-tuning the value for each problem or having an adaptive selection method could benefit the performances.  

\subsection{PT-ME Algorithm}

Algorithm~\ref{alg:main_alg} details PT-ME's pseudo-code\footnote[1]{Source code: \url{https://zenodo.org/doi/10.5281/zenodo.10926438}}. The two major changes with MT-ME~\cite{Multi-Task_MAP-Elites} are highlighted in colors and summarized to (1) a uniform sampling at each iteration of a new task parameter on which to evaluate a candidate solution and (2) a new variation operator using local linear regression used for 50\% of the iterations.



\begin{algorithm} 
    \small
    \caption{Parametric-Task MAP-Elites\\ Contributions: \colorbox{light-green}{Parametric-Task} \colorbox{light-blue}{New variation operator}}\label{alg:main_alg}
\begin{algorithmic}[1]

\State{} \highlightSection{\textbf{Parameters:}}

\State{} $d_x$: dimension of the solution space $\mathcal{X}={[0,1]}^{d_x}$ 
\State{} $d_{\theta}$: dimension of the task parameter space $\Theta={[0,1]}^{d_{\theta}}$ 
\State{} $\texttt{fitness}: \mathcal{X}\times\Theta \rightarrow \mathbb{R}$ 

\State{} \highlightSection{ \textbf{Hyper-parameters:}}

\State{} $B$: budget of evaluations (e.g., $100\,000$)
\State{} $N$: number of cells (e.g., $200$)
\State{} $S$: possible tournament sizes (e.g., $[1, 5, 10, 50, 100, 500 ]$)
\State{} $\sigma_{SBX}$: mutation factor for SBX~\cite{SBX} (e.g., $10.0$)  
\State{} \HLreg{$\sigma_{reg}$: mutation factor for our new variation operator (e.g., $1.0$)  }

\State{} \highlightSection{\textbf{Initialization:}}
\State{} $E \leftarrow \emptyset$ \Comment{$E$: to store all the evaluations}
\State{} $C \leftarrow$ $N$ random centroids using CVT \Comment{Divise $\Theta$ into cells}\label{lst:line:_init_CVT}
\State{} \HLreg{$D \leftarrow \operatorname{DelaunayTriangulation}(C)$}  \Comment{For computing adjacency}\label{lst:line:_init_delaunay}
\State{} \HLpar{$kdtree \leftarrow \operatorname{KDTree}(C)$} \Comment{To quickly find the cell of a task}\label{lst:line:_init_KDtree}
\State{} $A \leftarrow \emptyset$ \Comment{$A$: Archive}\label{lst:line:_init_archive_start}
\For{$\theta_{c}$ \texttt{in} $C$} \Comment{one for each cell}
    \State{} $A[\theta_{c}].\theta = \theta_{c}$   \Comment{$\theta$: the task parameter}
    \State{} $A[\theta_{c}].x = \texttt{random.uniform}(0, 1, )$  \Comment{$x:$ the solution}\label{lst:line:_init_random_solution}
    \State{} $A[\theta_{c}].f = \texttt{fitness}(A[\theta_{c}].x, A[\theta_{c}].\theta)$ \Comment{$f$: the fitness}
    \State{} \HLreg{$A[\theta_{c}].adj = D[\theta_{c}]$}  \Comment{$adj$: the adjacent cells}\label{lst:line:_init_archive_end}
\EndFor{} 

\State{} \highlightSection{\textbf{Main loop:}}
\State{} $s \leftarrow \operatorname{random\_in\_list(S)}$   \Comment{$s$: tournament size}
\State{} $selected \leftarrow \operatorname{zeros}(\operatorname{len}(S))$  \Comment{Counter of selection for each size}
\State{} $successes \leftarrow \operatorname{zeros}(\operatorname{len}(S))$  \Comment{Counter of successes for each size}
\For{$\_=0$ \texttt{to} $B-N$ } \Comment{We already consumed $N$ evaluations}
    \If{\HLreg{$\operatorname{random}() \le 0.5$}}  \Comment{Use SBX with tournament (50\%)}
        \State{} $p1, p2 \leftarrow$ two elites randomly selected from $A$\label{lst:line:_sbx_select_parents}
        \State{} \HLpar{$\theta_{1:s} \leftarrow \operatorname{random.uniform}(0, 1, (s, d_{\theta}))$} \Comment{Candidate tasks}\label{lst:line:_sbx_tasks_cnadidates}
        \State{} $selected[s] \leftarrow selected[s] + 1$ \Comment{Update the bandit}
        \State{} $\theta \leftarrow \operatorname{closest}(\theta_{1:s}, p1.\theta)$  \Comment{Task parameters tournament}\label{lst:line:_sbx_tournament}
        \State{} \HLpar{$\theta_c \leftarrow \texttt{kdtree.find}(\theta)$}\Comment{Find the corresponding cell}\label{lst:line:_sbx_corresponding_cell}
        \State{} $x \leftarrow \operatorname{SBX}(p1,p2,\sigma_{SBX})$  \Comment{Mutation \& Cross-over}\label{lst:line:_sbx_generate_offspring}
    \Else{} \Comment{Use new variation operator: Local Linear Regression (50\%) }
        \State{} \HLpar{$\theta \leftarrow \operatorname{random.uniform}(0, 1, d_{\theta})$}\label{lst:line:_llr_sample_task}
        \State{} \HLpar{$\theta_c \leftarrow \texttt{kdtree.find}(\theta)$}\Comment{Find the corresponding cell}\label{lst:line:_llr_corresponding_cell}
        \State{} \HLreg{$\boldsymbol{\theta} \leftarrow \{A[\theta_i].\theta\}_{\theta_{i} \in A[\theta_c].adj}$}\label{lst:line:_llr_adjacent_parameters}
        \State{} \HLreg{$\boldsymbol{x} \leftarrow \{A[\theta_i].x\}_{\theta_i \in A[\theta_c].adj}$}\label{lst:line:_llr_adjacent_solutions}
        \State{} \HLreg{$M \leftarrow ({\boldsymbol{\theta}}^\mathsf{T} \boldsymbol{\theta})^{-1} \boldsymbol{\theta}^\mathsf{T} \boldsymbol{x}$}\Comment{Linear least squares}\label{lst:line:_llr_lls}
        \State{} \HLreg{$x \leftarrow M \cdot \theta + \sigma_{reg} \cdot \mathcal{N}(0, \operatorname{variance}(\boldsymbol{x})$)}  \Comment{Regression $+$ noise}\label{lst:line:_llr_regression}
        \EndIf{}
        \State{} $f \leftarrow \texttt{fitness}(x, \theta)$ \Comment{Evaluate}\label{lst:line:_evaluation}
        \State{} \HLpar{$E.\operatorname{append}((\theta, x, f))$} \Comment{Store the evaluation}\label{lst:line:_evaluations}
    \If{ $f$ $\ge$ $A[\theta_c].f$}  \Comment{Update the archive}\label{lst:line:_achive_update_start}
        \State{} $A[\theta_c].\theta = \theta$
        \State{} $A[\theta_c].x = x$
        \State{} $A[\theta_c].f = f$\label{lst:line:_achive_update_end}
        \If{tournament was used}\Comment{Update the bandit}\label{lst:line:_bandit_update_start}
            \State{} $successes[s] \leftarrow successes[s] + 1$\label{lst:line:_bandit_update_end}
        \EndIf{}
    \EndIf{}
    \If{tournament was used} \Comment{Update the tournament size}\label{lst:line:_s_update_start}
        \State{} $s \leftarrow S\left[\underset{j}{\operatorname{argmax}} \biggl(\tfrac{successes[j]}{selected[j]}+\sqrt{\tfrac{2\ln(\operatorname{sum}(selected))}{selected[j]}}\biggr)\right]$\label{lst:line:_s_update_end}
    \EndIf{}
\EndFor{}
\State{} \Return{} $E$

\end{algorithmic}
\end{algorithm}

\subsection{Distillation}
\label{sec:_distillation}

\begin{algorithm} 
    \small
    \caption{Compute an archive with a new resolution}\label{alg:re_archive}
\begin{algorithmic}[1]

\State{} \highlightSection{\textbf{Parameters:}}

\State{} $d_{\theta}$: dimension of the task parameter space $\Theta={[0,1]}^{d_{\theta}}$ 
\State{} $E$: list of evaluations $(\theta, x, f)$ 
\State{} $N$: number of cells in the new archive 

\State{} \highlightSection{\textbf{Initialization:}}
\State{} $C \leftarrow$ $N$ random centroids using CVT \label{lst:line:_ra_init_CVT}
\State{} $kdtree \leftarrow \operatorname{KDTree}(C)$ 
\State{} $A \leftarrow \{ \theta_c: \{\theta:None, x:None, f:0\} \texttt{ for } \theta_c \texttt{ in }  C\}$\label{lst:line:_ra_init_archive}

\State{} \highlightSection{\textbf{Main loop:}}
\For{$(\theta, x, f)$ \texttt{in} $E$} \Comment{Compute elites}\label{lst:line:_ra_achive_update_start}
    \State{} $\theta_c \leftarrow \texttt{kdtree.find}(\theta)$\Comment{Find the corresponding cell}
    \If{ $f$ $\ge$ $A[\theta_c].f$} \Comment{We suppose $f$ normalized so that $f\ge0$ }
        \State{} $A[\theta_c] = (\theta, x, f)$\label{lst:line:_ra_achive_update_end}
    \EndIf{}
\EndFor{}

\State{} \Return{} $A$

\end{algorithmic}
\end{algorithm}

PT-ME asymptotically covers the whole task parameter space by visiting a new task at each iteration.
To solve the parametric-task optimization problem with a finite budget of evaluations and return a solution for each task parameter, we propose to learn an approximation $\hat{G}$ of the function $G:\theta \mapsto x^*_{\theta}$. To do so, Alg.~\ref{alg:re_archive} first computes an archive $A_{n}$ of elites for a different resolution $n$ using the dataset of evaluations returned by PT-ME\@. We then distill $A_{n}$'s elites into a multi-layer perceptron (with the same architecture as PPO's stable-baseline~\cite{PPO, stable-baselines3}, i.e., two hidden layers of 64 neurons) using a supervised regression training with the MSE loss.

\section{Experiments} 
\label{sec:_experiments}

\begin{figure*}
    \centering
    \includegraphics[width=1.\linewidth]{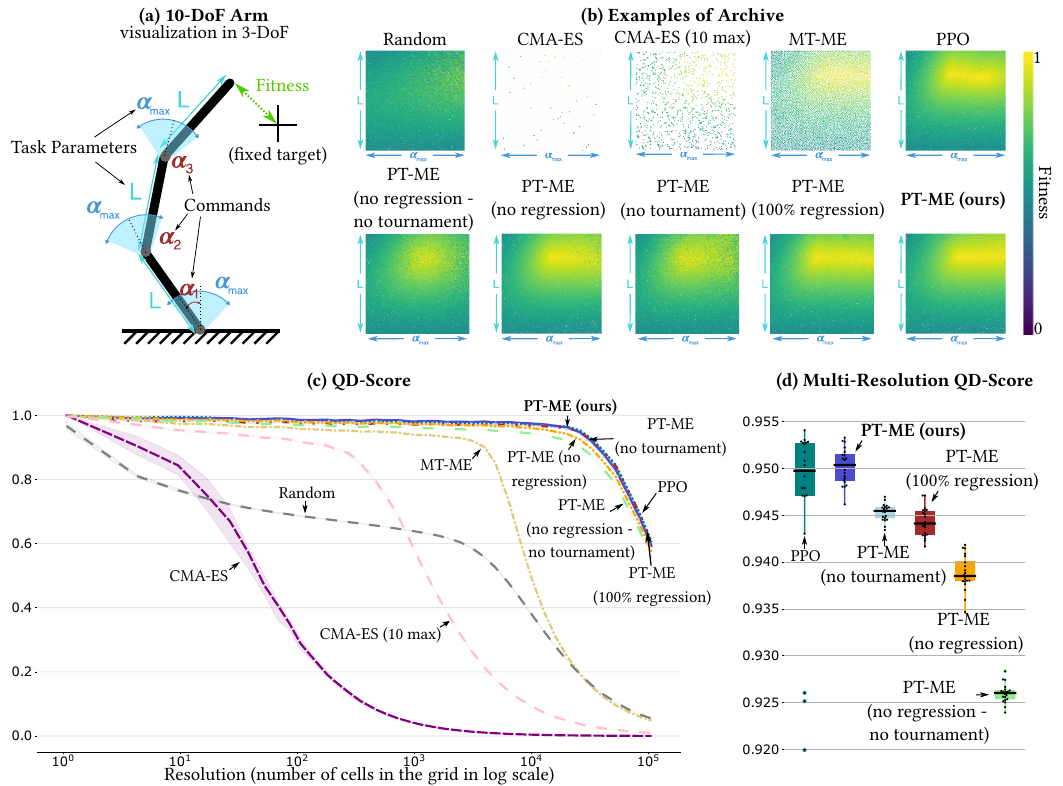}
    \caption{(a) Schematic view of the 10-DoF Arm problem. (b) Archive examples extracted from one run. (c) QD-Score for different resolutions (line $=$ median of 20 replications and shaded area $=$ first and third quantiles). (d) Multi-Resolution QD-Score.}\label{fig:main_comparison_arm}
\end{figure*}

\begin{figure*}
    \centering
    \includegraphics[width=1.\linewidth]{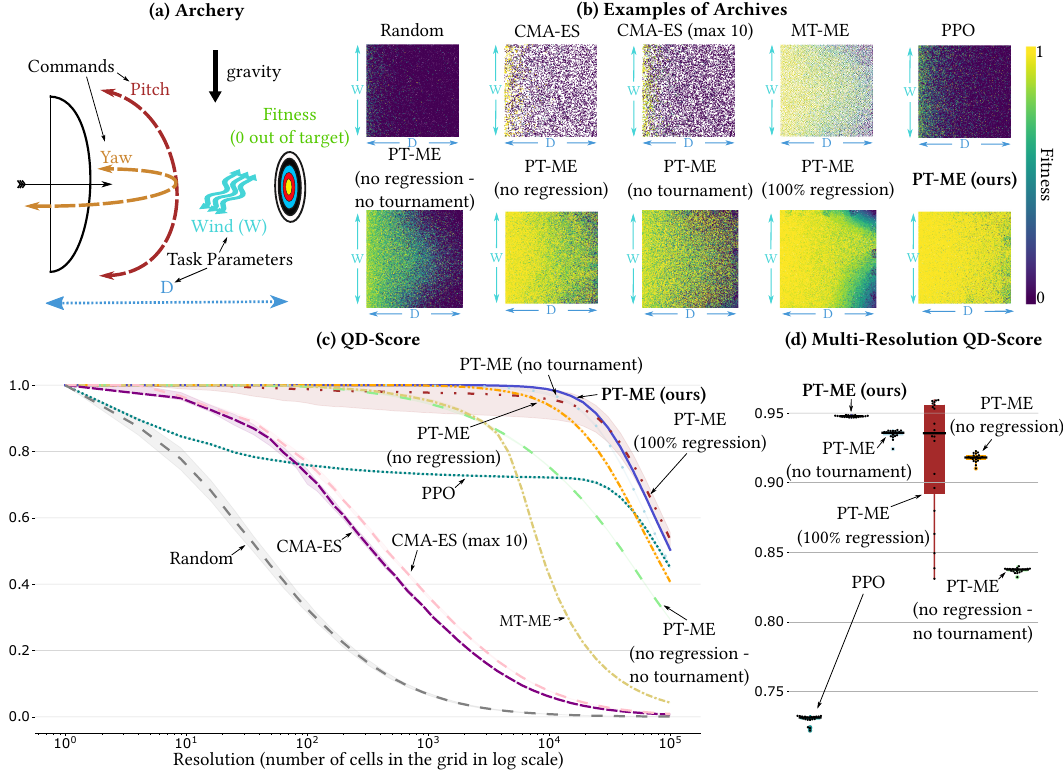}
    \caption{(a) Schematic view of the Archery problem. (b) Archive examples extracted from one run. (c) QD-Score for different resolutions (line $=$ median of 20 replications and shaded area $=$ first and third quantiles). (d) Multi-Resolution QD-Score.}\label{fig:main_comparison_archery}
\end{figure*}

\subsection{Considered Domains}
\label{sec:_Considered_Problems}
\subsubsection{10-DoF Arm} This toy problem is presented in MT-ME~\cite{Multi-Task_MAP-Elites} and reused in~\cite{EMT_planar_robotic_arm_anomaly_detection, many-tasks_EME-BI}. It consists in finding the joint positions for different 10-DoF arm morphologies that put the end effector as close to a fixed target (Fig.~\ref{fig:main_comparison_arm}.a). 

The task parameter space $\Theta$ corresponds to the maximal angular position $\alpha_{\max}$ and the length $L$ of each joint. The solution space $\mathcal{X}$ corresponds to the angular positions of each joint (normalized by the maximal angular position $\alpha_{\max}$). The fitness $f$ corresponds to the Euclidean distance between the end effector at position $pos_{ee}$ and the fixed target at position $pos_{tar}$. To get a maximization problem, we put this distance into a negative exponential: $f = \operatorname{\exp}(-||pos_{ee}-pos_{tar}||^2)$ so that it is bounded by 0 and 1.

\subsubsection{Archery} This toy problem corresponds to shooting an arrow at a range target and computing the associated score (Fig.~\ref{fig:main_comparison_archery}.a). 

The task parameter space $\Theta$ corresponds to the distance $D$ to the target (between $5$m and $40$m), and the strength and orientation of horizontal wind $W$, simulated by an additional constant acceleration (between $-10\mathrm{m}.\mathrm{s}^{-2}$ and $10\mathrm{m}.\mathrm{s}^{-2}$). The solution space $\mathcal{X}$ corresponds to the $yaw$ and $pitch$ of the arrow (between $-\tfrac{\pi}{12}$rad and $ \tfrac{\pi}{12}$rad) with a constant velocity $v$ of $70\mathrm{m}.\mathrm{s}^{-1}$. The fitness $f$ corresponds to the traditional archery scoring normalized between 0 and 1. More formally, we compute the velocity $\vec{v} = \left(\begin{smallmatrix}-\sin(yaw) \\ \cos(yaw) \cos(pitch) \\ \cos(yaw) \sin(pitch)\end{smallmatrix}\right)$, deduce the time of impact $t = \tfrac{D}{v_y}$, compute the distance to the target $d=||\tfrac{1}{2}(-g\vec{z}+W\vec{x})t^2+\vec{v}t||^2$ and $f=\max(0, int(10-\tfrac{d}{0.061}))/10$.


\subsubsection{Door-Pulling} A simulated humanoid robot opens a door by pulling a vertical handle (Fig.~\ref{fig:main_comparison_talos}.a). 

The task parameter space $\Theta$ corresponds to the position of the door relative to the robot (between $0.8$m and $1.2$m in front of the robot and $0$m and $-0.5$m laterally) and its orientation (between $\tfrac{3\pi}{8}$rad and $\tfrac{5\pi}{8}$rad). The fitness $f$ corresponds to the angle of the door after pulling. This fitness is sparse as the door angle stays null if the robot does not touch the handle. The solution space $\mathcal{X}$ corresponds to the following top-level command sent to a whole-body controller (WBC):
\begin{itemize}
    \item the gripper's translation from the handle of the door (3DoF);
    \item the gripper's orientation with constant pitch (2DoF);
    \item the gripper's translation during pulling (3DoF);
    \item the gripper's vertical rotation during pulling (1DoF).
\end{itemize}
The WBC solves at $1\,000$ Hz a quadratic programming problem that optimizes the robot joint positions to satisfy the quasi-static dynamics while trying to satisfy the top-level commands~\cite{dalin_whole-body_2021}.  

Door-Pulling is more similar to a DRL problem, where a policy would be trained to pull the door open from a random initial state. Instead, we explicitly optimize to find top-level commands that pull the door open for every possible configuration of the door with the goal of generalizing those solutions to the whole space.

\subsection{Methodology}

\subsubsection{Baselines}

\begin{itemize}
    \item \textbf{Random} uniformly samples the task parameter $\theta$ and the candidate solution $x$ (``How hard is the problem?'');
    \item \textbf{CMA-ES} iterates CMA-ES~\cite{CMA-ES} on a random task parameter till it reaches a stopping criterium, then switches to another till it reaches the budget $B$ (``How good is CMA-ES?'');
    \item \textbf{CMA-ES (max 10)} does the same but enforces a maximum of 10 iterations per task parameter before switching to another (``Does CMA-ES gain from seeing more different tasks?'');
    \item \textbf{PPO} runs the stable-baseline~\cite{stable-baselines3} PPO~\cite{PPO} with a random task parameter as initial state, the solution $x$ as action, and the fitness $f$ as reward (``Can DRL solve the problem?'');
    \item \textbf{MT-ME} runs MT-ME~\cite{Multi-Task_MAP-Elites} with $5\,000$ fixed tasks sampled with CVT~\cite{CVT} (``Do we need a new algorithm?'').  
\end{itemize}

\subsubsection{PT-ME variants}

\begin{itemize}
    \item \textbf{PT-ME (ours)} uses Alg.~\ref{alg:main_alg};
    \item \textbf{PT-ME (no regression)}\@ uses only SBX with tournament (``Is our new variation operator necessary?'');
    \item \textbf{PT-ME (100\% regression)} uses only the local linear regression (``Is our new variation operator sufficient?'');
    \item \textbf{PT-ME (no regression - no tournament)} uses only SBX with no tournament (``Is the tournament useful?'');
    \item \textbf{PT-ME (no tournament)} uses both variation operators with no tournament (``Is the tournament useful with the addition of our new variation operator?'').
\end{itemize}

\subsubsection{Evaluations and Measures}
\label{sec:_evaluations_and_measures}
After a budget $B$ of $100\,000$ evaluations, we want to evaluate both the coverage and the quality of the solutions with the idea that a perfect algorithm should find optimal solutions evenly spread on the task parameter space. 
We use the QD-Score introduced in~\cite{MAP-Elites} for a finite set of tasks: 
\[\text{QD-Score}(A) = \underset{(\theta, x, f)_{\theta_c}\in A}{\sum} f \] where $(\theta, x, f)_{\theta_c}$ is the elite in the cell of centroid $\theta_c$. 

As PT-ME stores all the evaluations $E$ from $B$ different tasks (\texttt{L}.\ref{lst:line:_evaluations}), we recompute the elites for different resolutions, ranging from one cell (the best-found solution on all tasks) up to $B$ cells evenly spread on the task parameter space $\Theta$ (we cannot fill more cells than the budget). We attribute the lowest fitness to each empty cell, meaning that looking at the QD-Score for the highest resolution does not give a complete idea of the algorithm's performance. To simplify the comparison, we introduce the Multi-Resolution QD-Score (MR-QD-Score) as the mean QD-Score for archives recomputed using Alg.~\ref{alg:re_archive} with a range of resolutions from 1 to $B$ cells. More formally: 
\begin{equation}
    \text{MR-QD-Score}(E, N_{1:n}) = \dfrac{1}{n} \overset{n}{\underset{i=1}{\sum}} \text{QD-Score}(\textbf{Alg.~\ref{alg:re_archive}}(E, N_i))
\end{equation}
where $N_{1:n}$ are different resolutions. We used $n=50$ resolutions evenly spread in logspace between 1 and $100\,000$ cells.

To ease the comparison between problems, we set the fitness minimal value to $0$ and its maximal value to $1$. It is already the case for Archery (i.e., the arrow can reach the center of the target, $f=1$, or miss the target, $f=0$). For the 10-DoF Arm, the fitness is bounded by $0$ and $1$. However, each task has different bounds (e.g., the arm can be too short to reach the target, leading to a maximal fitness strictly inferior to $1$). Knowing the bounds of the fitness functions for each task parameter is not trivial. We ran CMA-ES~\cite{CMA-ES} on each cell for the highest resolution to estimate them. 

All statistical tests are performed with a Mann–Whitney U test and the Bonferroni correction.

\subsection{Solutions Evaluations}

\subsubsection{10-DoF Arm}

Fig.~\ref{fig:main_comparison_arm} shows the results in the 10-DoF Arm problem of all methods with 20 replications each. 

\textbf{CMA-ES} allocates too many evaluations for the same task, which prevents it from covering the space and leads to a lower QD-Score than \textbf{Random} when there are more than 20 cells to fill. Bounding the number of iterations to 10 allows \textbf{CMA-ES (10 max)} to fill more cells with a better solution than \textbf{Random}, but it worsens after $1\,000$ cells. \textbf{MT-ME} finds solutions with quality similar to the best methods and outperforms \textbf{CMA-ES} and \textbf{CMA-ES (10 max)}, showing that solving all the tasks together is more efficient than solving them separately. However, due to its fixed set of tasks, it cannot fill more cells in higher resolutions, significantly decreasing its QD-Score after $5\,000$ cells. \textbf{PPO} and all variants of \textbf{PT-ME} have a similar QD-Score profile, succeeding in filling most of the cells with reasonable solutions, even in high resolutions. \textbf{PT-ME} 
and \textbf{PPO} 
have the highest MR-QD-Score and are not significantly different. 

\textbf{PT-ME}'s MR-QD-Score is significantly higher ($\text{p-value}$$<$$0.001$) than all the ablations, showing that the algorithm needs the local linear regression, SBX with a bandit-adapted tournament, and the parametric-task reformulation. Both \textbf{PT-ME (no tournament)} 
and \textbf{PT-ME (no regression - no tournament)} 
are significantly worse ($\text{p-value}$$<$$0.001$) than \textbf{PT-ME} 
and \textbf{PT-ME (no regression)}
, showing that the tournament is beneficial.

Fig.~\ref{fig:main_comparison_arm}.b shows archives for the different methods. \textbf{CMA-ES} and \textbf{MT-ME} have sparser archives as they allocate several evaluations per task parameter. \textbf{Random} fills all task parameter space but does not find solutions as good as \textbf{PPO} and \textbf{PT-ME} and its ablations.

\subsubsection{Archery}

We then compare all methods on the Archery problem (Fig.~\ref{fig:main_comparison_archery}.c) with 20 replications for each. Compared to 10-DoF Arm, where each candidate solution achieves a positive fitness, for Archery, most (i.e., approximately 95\%) of the solution space leads to a fitness of $0$, leading to \textbf{Random} having the worst performance. 

\textbf{CMA-ES} and \textbf{CMA-ES (10 max)} have similar QD-Score profiles because either the first generation of candidate solutions has a null fitness and the algorithm stops, or one of them reaches the target by chance, and CMA-ES quickly finds the optimal solution.

\textbf{MT-ME} outperforms \textbf{CMA-ES} and \textbf{CMA-ES (10 max)}, showing that solving all the tasks together is more efficient than solving them separately. For resolutions inferior to $10\,000$ cells, \textbf{MT-ME} is better than \textbf{PPO}, which fails in this problem due to its poor exploration with a sparse reward.

\textbf{PT-ME} 
is significantly better ($\text{p-value}$$<$$0.001$) than all the other methods except \textbf{PT-ME (100\% regression)} 
whose variance is one order of magnitude larger than all the other methods. As it is a pure exploitation method relying on ``luck'' to explore all space, it can easily get stuck with solutions that never expand the solved region.

Fig.~\ref{fig:main_comparison_archery}.b shows archives extracted from the executions of the different methods. \textbf{CMA-ES}, \textbf{CMA-ES (10 max)}, and \textbf{PPO} only solve the easiest tasks (i.e., where the target is the closest). \textbf{MT-ME} and PT-ME's ablations have a similar profile with worse solutions where the wind is the strongest and the target is the farthest. 

\subsubsection{Door-Pulling}

\begin{figure}
    \centering
    \includegraphics[width=1.\linewidth]{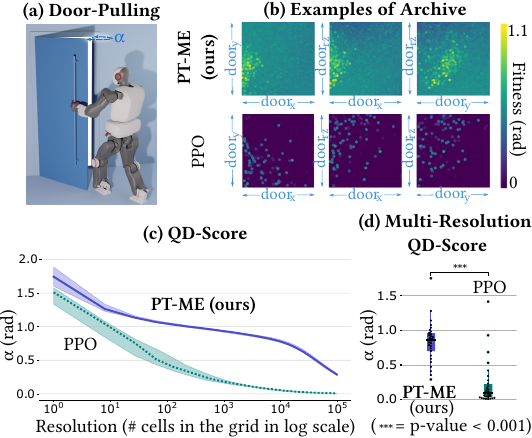}
    \caption{(a) Door-Pulling visualization. (b) Examples of the archive found by \textbf{PT-ME} and \textbf{PPO}~\cite{PPO}. (c) The QD-Score of the two methods for different archive resolutions (the line is the median of 10 replications, and the shaded area is between the first and third quantiles). (d) The Multi-Resolution QD-Score of the two methods for $10$ replications.}\label{fig:main_comparison_talos}
\end{figure}


Due to the computing cost of running the humanoid simulation (in the order of the dozens of seconds compared to the order of the millisecond for 10-DoF Arm and Archery), we only compared the result of \textbf{PT-ME} and \textbf{PPO}~\cite{PPO} on ten replications. The result (Fig.~\ref{fig:main_comparison_talos}) shows that our method significantly outperforms \textbf{PPO} ($\text{p-value}$$<$$0.001$), which can be explained again by the sparse fitness and \textbf{PPO}'s poor exploration abilities.



\subsection{Distillation Evaluations}
\begin{figure}
    \centering
    \includegraphics[width=1.\linewidth]{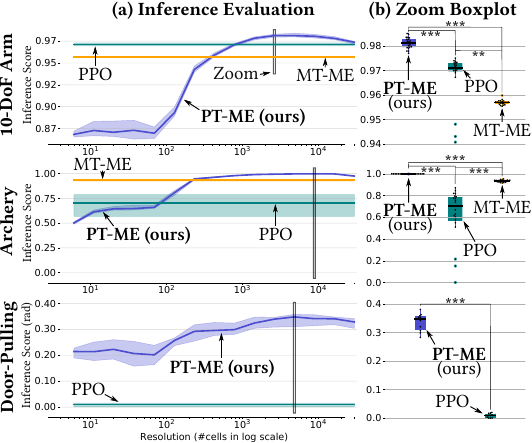}
    \caption{(a) Inference score, i.e., mean fitness over $10\,000$ ($1\,000$ for Door-Pulling) evenly spread new tasks of \textbf{PT-ME} distillation, \textbf{MT-ME} distillation, and \textbf{PPO} for the three parametric-task optimization problems (the line is the median of 20 (10 for Door-Pulling) replications, and the shaded area is between the first and third quantiles). (b) Box plots with \textbf{PT-ME}'s best resolution ( $*$$*$$*$$=$$\text{p-value}$$<$$0.001$, $*$$*$$=$$\text{p-value}$$<$$0.01$). }\label{fig:inference_comparison}

\end{figure}




We compare \textbf{MT-ME} (for one resolution) and \textbf{PT-ME} (for different resolutions) with distillation and \textbf{PPO} inference capacities by evaluating the fitness of their proposed solution for $10\,000$ ($1\,000$ for Door-Pulling) new tasks sampled with CVT~\cite{CVT} (Fig.~\ref{fig:inference_comparison}). \textbf{PT-ME} significantly outperforms \textbf{PPO} on all the three problems ($\text{p-value}$$<$$0.001$).  

\textbf{PT-ME} inference score first increases as the resolution of the archive increases, showing the need for a large dataset for distillation. The decrease in inference score for large resolutions (i.e., $\ge 10\,000$ cells) is due to the elites' decreases in quality as the selection pressure decreases. Running the algorithm for longer would increase the maximal resolutions with high-quality elites, even though for Archery, \textbf{PT-ME} reaches nearly a perfect score (the lowest run had an inference score of $0.997$).  For the two toy problems, \textbf{PT-ME} outperforms \textbf{MT-ME} for resolutions superior to 500 cells and reaches nearly perfect scores, showing that it covers the space with better solutions and solves the parametric-task optimization problem. For Door-Pulling, the inference score is significantly lower than the MR-QD-Score due to the chaotic behavior of the contact dynamics of our simulator. 

\section{Discussion and Conclusion}
PT-ME finds high-performing solutions for as many tasks as its budget allows it, making it easy to generalize and distill those solutions into a function approximation that effectively proposes a solution for each task parameter. Its inference significantly outperforms the DRL method PPO~\cite{PPO} on all three problems.

Can PT-ME be used for large task parameter spaces? PT-ME discretizes the task space using CVT~\cite{CVT}, which scales up to large dimensions~\cite{fast_cvt}. However, the archive needs more cells to keep a high resolution in large dimensions, which can slow down the convergence. Our new variation operator uses a local linear regression, which could be affected by a larger task space (i.e., the linear model requires the inversion of a matrix). Future work could study other supervised methods (e.g., neural networks) to perform regression.

Can \ourMethodAccronyme{} be used for large solution spaces? Future work could study the integration of two variation operators to leverage gradient information for scaling up MAP-Elites algorithms to large solution spaces (e.g., a policy neural network): leveraging differentiable problems by computing a gradient from the fitness function~\cite{MEGA} (which requires differentiable problems) and training a differentiable critic and exploits it~\cite{PGA-ME} (applicable to black-box problems). Such gradient-based variation operators could be incorporated into our parametric-task framework. 

\begin{acks}
This project is supported by the CPER SCIARAT, the CPER CyberEntreprises, the Direction General de l’Armement (convention Inria-DGA ``humanoïde résilient''), the Creativ’Lab platform of Inria/LORIA, the EurROBIN Horizon project (grant number 101070596),
the Agence de l'Innovation de Défense (AID), and
the ANR in the France 2030 program through project PEPR O2R AS3 (ANR-22-EXOD-007).
\end{acks}

\bibliographystyle{ACM-Reference-Format}
\bibliography{biblio}

\end{document}